\documentclass[11pt]{article}

\usepackage[preprint]{acl}

\usepackage{times}
\usepackage{latexsym}
\usepackage{amsmath}
\usepackage{hyperref}

\usepackage{multirow,xcolor,colortbl}
\usepackage{makecell}
\usepackage{adjustbox}
\usepackage{tikz}
\usepackage{tabularx}
\usepackage{wrapfig}

\usepackage{booktabs}   

\usepackage[capitalize]{cleveref}
\crefname{section}{Sec.}{Secs.}
\Crefname{section}{Section}{Sections}
\Crefname{table}{Table}{Tables}
\crefname{table}{Tab.}{Tabs.}

\definecolor{stepcolor}{RGB}{140, 140, 250} 
\definecolor{skyblue}{RGB}{56, 166, 235}
\definecolor{softpurple}{RGB}{147, 112, 219}
\definecolor{softgreen}{RGB}{60, 179, 113}
\newcommand{\stepcircle}[1]{%
  \tikz[baseline=(char.base)]{%
    \node[shape=circle, fill=stepcolor, text=white, inner sep=0.8pt, minimum size=0.5em, font=\sffamily\bfseries] (char) {#1};%
  }%
}

\usepackage[T1]{fontenc}

\usepackage[utf8]{inputenc}

\usepackage{microtype}

\usepackage{inconsolata}

\usepackage{graphicx}
\newcommand{\eg}{\emph{e.g.}}

\usepackage{enumitem}

\usepackage[most]{tcolorbox}
\usepackage{listings}
\usepackage{stfloats}

\definecolor{promptbg}{RGB}{232,246,232}   
\definecolor{promptbd}{RGB}{150,200,150}   
\lstdefinestyle{promptstyle}{
  basicstyle=\ttfamily\scriptsize,   
  breaklines=true,
  columns=fullflexible,
  showstringspaces=false,
  frame=none,
  xleftmargin=0pt,
  gobble=0
}

\title{Think, Look, and Revise: Inconsistency-Aware Visual Self-Correction in MLLMs}

\author{Yu Cheng \\
  University of Edinburgh \\
  \texttt{s2521923@ed.ac.uk} \\\And
  Arushi Goel\\
  NVIDIA Research  \\
  \texttt{arushig@nvidia.com}\\\And
  Hakan Bilen \\
  University of Edinburgh  \\
  \texttt{hbilen@ed.ac.uk} \\}

\begin{document}
\twocolumn[{%
    \renewcommand\twocolumn[1][]{#1}%
    \maketitle
    \centering
    \vspace{-2em}
    \includegraphics[width=\linewidth]{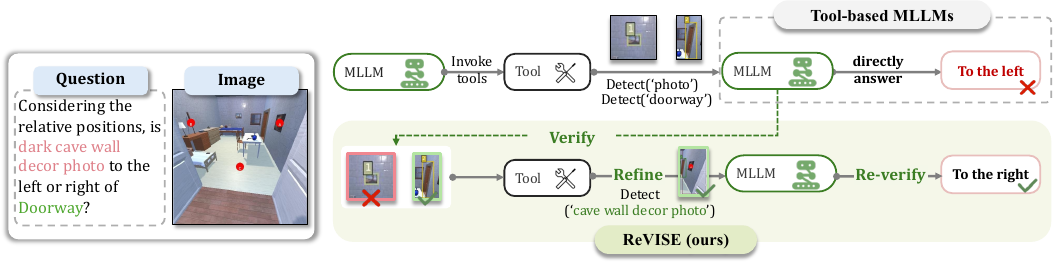}
    \captionof{figure}{\textbf{Overview of ReVISE (ours).} Conventional tool-augmented MLLMs (top) operate in an open-loop manner, blindly trusting intermediate tool outputs. If a tool misidentifies the correct visual target (\eg, detecting a generic ``photo'' instead of the intended ``dark cave wall decor photo''), the error propagates to the final answer. \textbf{ReVISE} (bottom) introduces an active self-correction loop: the model \textit{verifies} tool-derived evidence against the visual query, detects mismatches, \textit{refines} the tool query, and \textit{re-verifies} the updated evidence to obtain the accurate spatial relationship.}
    \label{fig:teaser}
    }
    ]

\begin{abstract}
Tool-augmented multimodal reasoning integrates external tools (e.g., object detection, depth estimation) into multimodal large language models (MLLMs) to address perceptual bottlenecks in complex visual tasks. However, existing approaches rarely verify tool outputs, limiting their ability to detect and recover from tool failures. We propose ReVISE, a framework that equips MLLMs with verification and dynamic error recovery for tool-augmented reasoning. ReVISE introduces (1) a curated training dataset that supervises reflective behaviors, enabling models to validate tool-derived evidence, reformulate queries when visual mismatches arise, and fall back to intrinsic grounding when external tools are unreliable; and (2) a reinforcement learning based targeted rewards that encourage internal reflection and penalize spatial misalignment. Experiments on several benchmarks demonstrate consistent improvements over existing methods, highlighting the importance of error detection and correction in tool-augmented multimodal reasoning.
\end{abstract}


\section{Introduction}
\label{sec:intro}

Multimodal large language models (MLLMs) \citep{liu2024sphinx, li2024llava, Liu_2024, DBLP:journals/corr/abs-2502-13923, DBLP:journals/corr/abs-2511-21631, DBLP:journals/corr/abs-2504-10479, DBLP:journals/corr/abs-2508-18265} are now widely used for vision tasks that require both perception and reasoning, such as grounding, counting, and spatial relation understanding. Recent open-source model families such as Qwen VL \citep{DBLP:journals/corr/abs-2502-13923, DBLP:journals/corr/abs-2511-21631}, and InternVL \citep{DBLP:journals/corr/abs-2504-10479, DBLP:journals/corr/abs-2508-18265} demonstrate strong performance across a broad range of benchmarks, benefiting from increasingly common post-training recipes that combine supervised fine-tuning (SFT) with reinforcement learning (RL). However, one critical issue is that reasoning chains remain purely textual even for multimodal inputs, which often limits visual interpretability due to the insufficient visual evidence at intermediate reasoning steps \citep{DBLP:journals/corr/abs-2505-15879}.

To address this limitation, tool-integrated multimodal reasoning augments MLLMs with callable external vision tools such as zooming, cropping, and object detection that return visual evidence, enabling the model to ground intermediate decisions. Recent open-source efforts \citep{DBLP:journals/corr/abs-2512-16918, DBLP:journals/corr/abs-2509-01656, DBLP:journals/corr/abs-2512-17312, hong2025deepeyesv2, Wang_2025, su2025openthinkimg} such as ReVPT \citep{DBLP:journals/corr/abs-2509-01656}, Thyme \citep{DBLP:journals/corr/abs-2508-11630}, and CodeDance \citep{DBLP:journals/corr/abs-2512-17312}, instantiate this paradigm with multi-step interaction traces, where the model alternates between generating intermediate rationales and invoking visual tools, then incorporates the returned observations as visual evidence to produce the final prediction. 

Despite their effectiveness, these methods primarily focus on \emph{how to use tools}, rather than \emph{how to evaluate and revise unreliable tool outputs}. Consequently, they exhibit an \emph{open-loop} behavior, treating intermediate tool outputs as correct and propagating them without verification. This prevents models from detecting inconsistencies between tool-derived evidence and the visual query, limiting their ability to recover from noisy or misaligned outputs. The issue stems from two gaps: SFT data largely lacks supervision for failure detection and correction, and RL objectives emphasize final answer accuracy without incentivizing verification or revision. Furthermore, over-reliance on tools weakens intrinsic visual reasoning, causing models to behave as tool callers rather than autonomous visual reasoners, especially when tools are unavailable or unreliable.

In this paper, we propose \textbf{ReVISE}, a novel framework designed to close the \emph{open-loop} in tool-augmented multimodal reasoning. Instead of blindly deferring to external tool outputs, 
ReVISE moves toward a \emph{verification-driven} and \emph{visually grounded} reasoning process, where tool outputs are validated against the model’s direct interpretation of the image.
We equip the MLLM with the active capability to critically evaluate intermediate visual observations, triggering dynamic self-correction when tool feedback is noisy, contradictory, or missing. 
As illustrated in \cref{fig:teaser}, while standard tool-augmented MLLMs suffer from reasoning failures when initial tool detections are imprecise or misaligned, ReVISE rejects noisy observations and triggers a refined tool execution to acquire the correct visual matching before generating the final answer by explicitly verifying the returned visual evidence against the multimodal context.

Specifically, we first construct a high-quality cold-start dataset sourced from SAT \citep{ray2024sat}, TACO \citep{Ma_2025}, and TallyQA \citep{Acharya_2019}. 
Our curated dataset features multi-turn tool-interaction trajectories that explicitly embed diverse verification and revision patterns. Building upon this foundation, we advance the model through Group Relative Policy Optimization (GRPO) \citep{DBLP:journals/corr/abs-2402-03300}-based RL method governed by a novel, correction-oriented reward mechanism. 

In summary, our contributions are as follows:
    \textbf{(i) Verification-driven tool-augmented reasoning.} We introduce a new paradigm that explicitly models tool-output verification and correction as learnable behaviors, addressing the open-loop limitation of existing tool-augmented MLLMs.
    \textbf{(ii) Reflection-supervised training for self-correction.} We construct a dataset with diverse multi-turn trajectories that explicitly supervise inconsistency detection, verification, and revision, enabling the model to learn when and how to distrust tool outputs.
    \textbf{(iii) Correction-oriented reinforcement learning.} We propose targeted reward signals, including a self-correction reward that encourages effective revision and a grounding reward that optimizes spatial refinement of bounding boxes.
    \textbf{(iv) Improved robustness in perception-heavy tasks.} Extensive experiments on Qwen2.5-VL \citep{DBLP:journals/corr/abs-2502-13923} show that ReVISE consistently improves performance on challenging benchmarks such as CountBench \citep{Paiss_2023}, CVBench \citep{Akula_2024}, and BLINK-HARD \citep{bigverdi2025perception}, demonstrating the effectiveness of explicit verification and correction.

\section{Related Work}\label{sec:related}
\textbf{Multimodal Large Language Models.}
Multimodal large language models (MLLMs) extend large language models with visual perception, enabling joint reasoning over images and text. Instruction-tuned systems such as the LLaVA family \citep{DBLP:conf/nips/LiuLWL23a, Liu_2024}, LLaVA-OneVision \citep{li2024llava}, and NVILA \citep{Liu_2025} demonstrate that scaling multimodal instruction data substantially improves performance across diverse visual benchmarks.

Most MLLMs follow a multi-stage training pipeline. After large-scale pretraining, models are adapted to multimodal tasks via SFT, which aligns model outputs with human instructions and reasoning formats. Building on this initialization, recent work explores post-training optimization techniques such as RL and preference-based learning to further improve reasoning capabilities \citep{DBLP:conf/nips/RafailovSMMEF23, DBLP:journals/corr/abs-2402-03300}. Methods including
Direct Preference Optimization (DPO) \citep{DBLP:conf/nips/RafailovSMMEF23} and Group Relative Policy Optimization (GRPO) \citep{DBLP:journals/corr/abs-2402-03300} provide efficient alternatives to traditional RL pipelines for optimizing reasoning behavior. Despite these advances, MLLMs still struggle with perception-heavy reasoning that requires precise visual grounding or verification of visual evidence, motivating approaches that augment them with external perceptual tools.

\noindent\textbf{Tool-Augmented Visual Reasoning.}
To address these perceptual limitations, several works integrate external vision tools into multimodal reasoning pipelines. Early compositional methods such as ViperGPT \citep{DBLP:conf/iccv/SurisMV23} and VisProg \citep{DBLP:conf/cvpr/GuptaK23} translate natural language queries into executable programs that invoke specialized vision modules via predefined prompts, enabling complex visual questions to be decomposed into sequences of tool operations.
More recent work \citep{DBLP:journals/corr/abs-2512-16918, DBLP:journals/corr/abs-2509-01656, DBLP:journals/corr/abs-2512-17312, hong2025deepeyesv2, Wang_2025, su2025openthinkimg} explores learning-based approaches that enable models to predict tool invocations automatically. Some methods \citep{DBLP:conf/cvpr/HuSLVHLKF24} distill tool-augmented pipelines into end-to-end models, while others \citep{DBLP:journals/corr/abs-2507-07998, DBLP:journals/corr/abs-2509-01656, DBLP:journals/corr/abs-2508-11630, DBLP:journals/corr/abs-2512-16918, DBLP:journals/corr/abs-2512-17312} apply RL to improve tool selection and reasoning policies. 
Although these approaches improve visual reasoning performance, they typically focus on using tools to obtain the correct answer without explicitly modeling how the reasoning process should adapt when tool evidence contradicts the model's initial prediction.

Recent methods such as CodeDance \citep{DBLP:journals/corr/abs-2512-17312} and AdaTooler-V \citep{DBLP:journals/corr/abs-2512-16918} explore iterative reasoning strategies that allow models to revisit visual evidence during inference. However, these approaches largely rely on inference-time heuristics \citep{DBLP:journals/corr/abs-2512-17312, DBLP:journals/corr/abs-2508-11630, DBLP:journals/corr/abs-2509-01656} and do not explicitly supervise correction-oriented reasoning during training.
In contrast, our work focuses on learning \emph{visual self-correction}. 
We construct training trajectories that encourage the model to revise its predictions when new visual evidence contradicts its initial hypothesis and introduce supervision signals that promote object-level grounding during reasoning. 
By explicitly modeling how reasoning evolves as additional visual evidence becomes available, our approach improves robustness in tool-augmented visual reasoning.

\section{Method}
\label{sec:method}

ReVISE is a verification-driven framework for tool-augmented multimodal reasoning. We first formulate tool-integrated reasoning as a sequential decision process, and then optimize the policy with rewards that encourage faithful reflection and accurate visual grounding. 
The construction of our reflection-rich supervised data is described separately in \cref{sec:data_curation}.
Here, we define \emph{verification} as the process of assessing the consistency between tool-derived observations, the visual query, and the model’s own perception, and triggering revision when mismatches are detected.

\subsection{Problem Formulation}
\label{subsec:problem_formulate}
In this paper, reasoning can be formulated as a multi-step inferential process in which the MLLM $\pi$ iteratively makes intermediate decisions and, when necessary, invokes external visual tools to gather evidence before producing a final answer. Given an input sample $(i, q)$, where $i$ denotes the visual input and $q$ the textual query, the reasoning procedure is commonly modeled as a think--execute--feedback trajectory.
At each step, the model conditions on the current query and its accumulated reasoning context, selects an action (either a tool invocation or a terminal answer), and receives an observation returned by the tool execution. 
Formally, it can be formulated as:
\begin{equation}
    \tau = ((s_1, a_1, o_1), \dots, (s_T, a_T, o_T))
\end{equation}
where $T$ is the maximum number of steps. 
The state $s_t = (i, q, h_t)$ encapsulates the initial visual input $i$, the textual query $q$, and the accumulated trajectory history $h_t = (a_1, o_1, \dots, a_{t-1}, o_{t-1})$ up to step $t$.
At each step, the model $\pi$ samples an action $a_t \sim \pi(\cdot \mid s_t)$ from a unified space $\mathcal{A}$ comprising visual tool invocations and a terminal answer. Executing a selected tool yields a spatial observation $o_t$ that updates the subsequent state to $s_{t+1}$ (with $o_t = \emptyset$ for non-tool actions). This sequential decision process iterates until the model emits the final answer or the step reaches the maximum step $T$.

\subsection{Cold-Start Supervised Fine-Tuning}
Given a set of curated reasoning trajectories $\mathcal{D} = \{ (i^{(n)}, q^{(n)}, \tau^{(n)}) \}_{n=1}^N$, we first perform supervised fine-tuning to teach the model the basic patterns of tool use, verification, and self-correction. 
Each trajectory  $\tau=\{(s_t,a_t,o_t)\}_{t=1}^{T}$ exemplifies a sequence of optimal actions and observations that lead to a correct answer, including cases where the model must detect and revise inconsistent evidence where $T$ is the maximum number of steps.
We optimize the MLLM parameters $\theta$ by next-token prediction over the action sequence:
\begin{equation}
\resizebox{0.78\linewidth}{!}{$
\begin{aligned}
\mathcal{L}_{\mathrm{SFT}}(\theta)
&= -\frac{1}{N}\sum_{n=1}^{N}\log \pi_{\theta}\!\left(\tau^{(n)} \mid i^{(n)}, q^{(n)}\right) \\
&= -\frac{1}{N}\sum_{n=1}^{N}\sum_{t=1}^{T^{(n)}} \log \pi_{\theta}\!\left(a^{(n)}_t \mid s^{(n)}_t\right)
\end{aligned}
$}
\end{equation}
By minimizing $\mathcal{L}_{\text{SFT}}$, the model acquires a strong cold-start capability for sequential visual tool use and self-correction.

\subsection{Policy Optimization via Reflection and Grounding Rewards}
\label{subsec:grpo}


Next, we further optimize the model policy using RL. 
To encourage the model to reflect on intermediate evidence and continuously improve its intrinsic visual grounding capability, we build on GRPO \citep{DBLP:journals/corr/abs-2402-03300} (detailed formulation of standard GRPO can be seen in Appendix~\ref{sec:grpo_fomulation}) and introduce reward components tailored to verification and spatial refinement.

Specifically, we define the task-specific reward $r_k$ as the sum of four sub-rewards: 

\noindent \textbf{Answer-Accuracy Reward} ($r_{\text{accuracy}}$) measures the correctness of the final answer. Concretely, we extract the prediction from the \textcolor{softpurple}{\texttt{<answer>}} \dots \textcolor{softpurple}{\texttt{</answer>}} block and verify it with a rule-based answer checker.

\noindent \textbf{Reasoning Format Reward} ($r_{\text{format}}$) ensures that the model follows the prescribed generation format during multi-turn inference. The model must enclose its internal reasoning within \textcolor{skyblue}{\texttt{<think>}} and \textcolor{skyblue}{\texttt{</think>}} tokens, and its tool invocations within \textcolor{softgreen}{\texttt{<tool\_call>}} and \textcolor{softgreen}{\texttt{</tool\_call>}} tokens, or produce a final response within \textcolor{softpurple}{\texttt{<answer>}} and \textcolor{softpurple}{\texttt{</answer>}}. The reward is set to $1$ if the output matches the expected format and $-1$ otherwise.

\noindent \textbf{Self-Correction Reward} ($r_{\text{correction}}$) encourages dynamic verification and revision of intermediate evidence. It is set to $1$ if the final answer is correct and the intermediate reasoning trace contains the predefined self-correction token \textcolor{orange}{\texttt{re-examine}}. It is set to $-1$ if the model produces this token but still arrives at an incorrect answer, and $0$ otherwise.
During training, this token is extracted via regular expression parsing. To maximize extraction reliability, the SFT data and formatting instructions explicitly require the model to emit this cue whenever it initiates an internal reflection phase.

\noindent \textbf{Grounding Reward} ($r_{\text{grounding}}$) improves the model's intrinsic grounding capability by requiring it to output object bounding boxes in the final answer block. These coordinates may originate either from explicit tool use or from the model's own spatial reasoning. 
First, we define a static intersection over union (IoU) reward as the mean IoU between the final predicted boxes and ground-truth boxes:
$r_{\text{IoU}} = \frac{1}{|\hat{B}|} \sum_{j=1}^{|\hat{B}|} \text{maxIoU}(\hat{B}_j, \{B\})$,
where $\text{maxIoU}(\hat{B}_j, \{B\})= \max_{k} \text{IoU}(\hat{B}_j, B_k)$. 
$|\hat{B}|$, $\hat{B}$ and $B$ denote the number of predicted boxes,the predicted and ground-truth boxes, respectively. 
Second, we introduce $r_{\text{IoU++}}$ that captures localization improvement over the reasoning trajectory:
\begin{equation}
\resizebox{0.80\linewidth}{!}{$
\begin{aligned}
r_{\text{IoU++}}
&= \frac{1}{|\hat{B}|}\sum_{j=1}^{|\hat{B}|} \max\!\Big( 0,\; \text{maxIoU}(\hat{B}^{(T)}_j, \{B\})\\
&- \text{maxIoU}(\hat{B}^{(0)}_j, \{B\}) \Big)
\end{aligned}
$}
\end{equation} where $\hat{B}^{(0)}$ denote the initial box prediction and $\hat{B}^{(T)}$ the final refined prediction.
$r_{\text{IoU++}}$ rewards only positive improvements in spatial alignment. If no bounding boxes are produced or updated during the trajectory, we set $r_{\text{IoU++}}=0$. We then sum both terms as
$r_{\text{grounding}} = r_{\text{IoU}} + r_{\text{IoU++}}$.


\section{Training Data Curation}
\label{sec:data_curation}

To teach ReVISE to verify tool outputs, revise inconsistent evidence, and fall back to intrinsic perception when needed, we construct a reflection-rich supervised dataset with diverse tool-use and tool-free reasoning trajectories.

\noindent\stepcircle{1} \textbf{Toolset Collection.} We assemble a specialized suite of visual perception tools designed to address the inherent perceptual bottlenecks of MLLMs. Inspired by prior tool-augmented methods \citep{DBLP:journals/corr/abs-2509-01656}, we integrate four representative visual operations, including an open-vocabulary object detector for box-level grounding, a ZoomIn crop-and-resize operator for fine-grained inspection, edge detection for boundary cues, and depth estimation for geometric ordering and occlusion reasoning. Implementation details are deferred to Appendix~\ref{sec:tool_details}.

\noindent\stepcircle{2} \textbf{Trajectory Design.} As mentioned above, tool-integrated reasoning is cast as a decision-making process where MLLMs must learn to reason step by step, dynamically deciding when to invoke external tools and how to interpret their feedback. However, merely training on tool invocation is insufficient, as reliable performance also requires strong intrinsic visual grounding and spatial reasoning, as well as verification-driven revision of intermediate feedback. The key challenge is learning to detect evidence mismatches between tool outputs and the queried visual concept, and revise its plan by either re-invoking tools with refined arguments or falling back to the MLLM's own visual inspection.

\begin{figure*}[htbp]
  \centering
  \vspace{-10pt}
  \includegraphics[width=\linewidth]{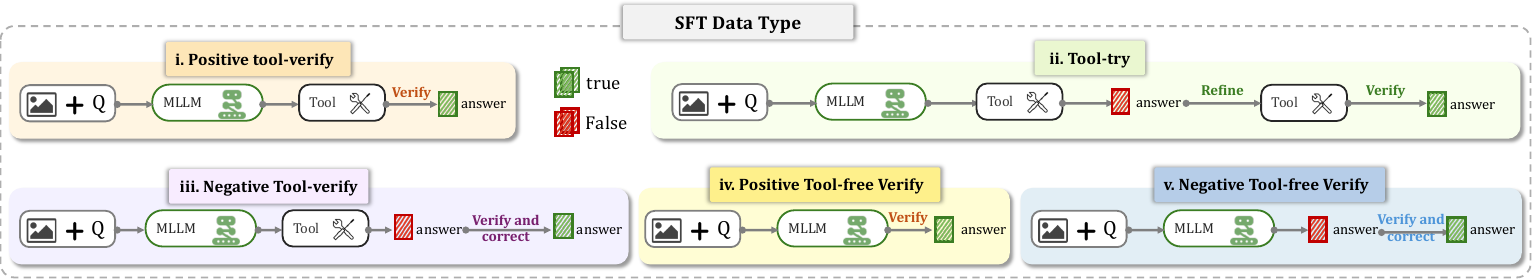}
  \caption{Five types of training trajectories to supervise verification and self-correction, illustrating the combinations of negative and positive verification with or without tool use. 
  }
  \label{fig:training_example}
\end{figure*}

\cref{fig:training_example} illustrates five trajectory types for cold-start supervision:
\textit{(i) Positive Tool-Verify}: the tool returns correct and query-aligned observations, and the MLLM explicitly verifies them and uses them to answer;
\textit{(ii) Tool-Retry}: the initial tool output is noisy or mismatched, prompting the MLLM to detect the inconsistency and re-invoke the tool with a revised query or select a different tool;
\textit{(iii) Negative Tool-Verify}: the tool fails (\eg, missing detections), and the MLLM verifies the failure and falls back to intrinsic perception;
\textit{(iv) Positive Tool-Free Verify}: the initial reasoning is correct and the MLLM confirms it without tool use; and
\textit{(v) Negative Tool-Free Verify}: the initial reasoning is incorrect and the MLLM performs verification and revision without tool use.

\noindent\stepcircle{3} \textbf{Trajectory Generation and Training.} We synthesize the reasoning traces using Gemini-2.5-pro \citep{comanici2025gemini}. Specifically, we apply specialized instructional prompts for each trajectory type. We then manually filter out generated trajectories that contain errors.

\section{Experiments}
\label{sec:experiments}
\subsection{Experiment Setup}\label{subsec:setup}

\textbf{Models.} We adopt the Qwen-VL series, specifically Qwen2.5-VL \citep{DBLP:journals/corr/abs-2502-13923} as our backbone model, and conduct experiments on both Qwen2.5-VL-3B and Qwen2.5-VL-7B. 
In our experiments, we initialize our model with the pre-trained weights from Qwen2.5-VL \citep{DBLP:journals/corr/abs-2502-13923} for the cold-start SFT stage, which is then further optimized using \cref{eq:rlobj}. More details of the training setting can be seen in Appendix~\ref{sec:additional_implement}.

\noindent \textbf{Dataset Construction.} 
Following ReVPT \citep{DBLP:journals/corr/abs-2509-01656}, we compile data from three diverse sources, the SAT dataset \citep{ray2024sat}, TACO \citep{Ma_2025} and TallyQA \citep{Acharya_2019}. 
SAT \citep{ray2024sat} is a synthetic visual question answering dataset designed to improve visual perception. 
TACO \citep{Ma_2025} provides multi-modal reasoning steps well-suited for tool-assisted responses, and TallyQA \citep{Acharya_2019} introduces complex visual counting and spatial reasoning challenges that require precise grounding. 
We filter the above data using Qwen2.5-VL-7B-Instruct \citep{DBLP:journals/corr/abs-2502-13923} and retain the samples it answers incorrectly to ensure that the samples are sufficiently difficult. Further details of applied prompts for tool-use trajectories and ground-truth bounding boxes (for objects referenced in the question-answer pairs) used for the grounding reward in \cref{subsec:grpo}, and data statistics are listed in Appendix~\ref{sec:curated_dataset}. 
To facilitate reproducibility and future research, we will publicly release our data and models.


For evaluation, we test our model across a diverse suite of multimodal benchmarks, including the complex counting like CountBench~\cite{Paiss_2023}, spatial reasoning tasks such as CVBench~\cite{Akula_2024}, BLINK~\cite{Fu_2024}, BLINK-HARD~\cite{bigverdi2025perception} and MMVP~\cite{Tong_2024}, comprehensive general multimodal evaluations such as MMSTAR~\cite{Chen_2024}, MMMU~\cite{Yue_2024}, and MMBench~\cite{Yue_2024} and math reasoning like MathVista~\cite{lu2023mathvista}.


\noindent \textbf{Baselines.} 
We compare our model against a diverse set of strong baselines, categorized into three groups. 
First, we evaluate leading \textbf{proprietary MLLMs}, including Gemini-2.0-Flash~\cite{comanici2025gemini} and GPT-4.1~\cite{achiam2023gpt}.
Second, we compare against strong \textbf{general open-sourced MLLMs}, specifically the Qwen2.5-VL-Instruct~\cite{DBLP:journals/corr/abs-2502-13923} series for 3B and 7B variants.
Third, we benchmark against \textbf{recent tool-augmented methods}, including TACO~\cite{Ma_2025} (based on Qwen2-VL-7B), ReVPT~\cite{DBLP:journals/corr/abs-2509-01656} (3B and 7B), Thyme-7B~\cite{DBLP:journals/corr/abs-2508-11630}, and the recently introduced CodeDance-7B~\cite{DBLP:journals/corr/abs-2512-17312}. 
Finally, to isolate the improvements brought by our proposed reasoning data and novel reward design, we introduce internal baselines by applying standard SFT data without tool and standard GRPO to the base Qwen2.5-VL models (denoted as Qwen2.5-VL-SFT and Qwen2.5-VL-SFT-GRPO, respectively)

\begin{table*}[!t]
\centering

\setlength{\tabcolsep}{3.5pt}
\renewcommand{\arraystretch}{1.12}

\begin{adjustbox}{width=\textwidth}
\begin{tabular}{l |c |c |c |c |c |c |c |c |c |c |c |c |c |c}
\toprule
\multirow{2}{*}{\textbf{Model}} &
\multirow{2}{*}{\textbf{Avg.}} &
\multirow{2}{*}{\textbf{CountBench}} &
\multicolumn{5}{c|}{\textbf{CV-Bench}} &
\multirow{2}{*}{\textbf{BLINK(sub)}} &
\multirow{2}{*}{\textbf{MMVP}} &
\multirow{2}{*}{\textbf{MMSTAR}} &
\multirow{2}{*}{\textbf{BLINK-HARD}} &
\multirow{2}{*}{\textbf{MMMU}} &
\multirow{2}{*}{\textbf{MMBench}} &
\multirow{2}{*}{\textbf{MathVista}} \\
\cmidrule(lr){4-8}
& & & Count & Relation & Depth & Distance & Avg. & & & & & & & \\
\midrule
\rowcolor{gray!15} Gemini-2.0-Flash~\cite{comanici2025gemini} & 77.35 & 87.43 & 71.95 & 86.92 & 87.50 & 82.17 & 81.50 & 76.37 & 79.34 & 69.40 & 68.28 & 71.70 & 89.05 & 73.10 \\
\rowcolor{gray!15} GPT-4.1~\cite{achiam2023gpt}         & 77.36 & 87.78 & 67.77 & 92.00 & 94.50 & 89.50 & 84.76 & 68.80 & 88.00 & 69.80 & 66.13 & 74.00 & 86.60 & 70.40 \\
\midrule
\rowcolor{purple!15} Qwen2.5-VL-3B-Instruct~\cite{DBLP:journals/corr/abs-2502-13923}  & 62.00 & 72.21 & 68.65 & 74.92 & 76.00 & 71.67 & 72.55 & 64.34 & 62.67 & 53.40 & 41.67 & \textbf{49.00} & 80.97 & \textbf{61.20} \\
Qwen2.5-VL-3B-SFT       & 63.10 & \underline{73.86} & 62.58 & 91.78 & \underline{87.34} & {\textbf{83.95}} & \underline{82.78} & 64.62 & 59.89 & 50.37 & 58.62 & 41.98 & 81.24 & 54.54 \\
Qwen2.5-VL-3B-SFT-GRPO      & 62.03 & 73.41 & 69.16 & 81.23 & 80.17 & 61.50 & 72.90 & 60.66 & \underline{69.00} & 46.60 & 51.61 & 45.44 & \underline{83.69} & 54.95 \\
ReVPT-3B~\cite{DBLP:journals/corr/abs-2509-01656}               &   \underline{66.77}    &   73.19    & \underline{70.43} & \underline{92.62} & \textbf{87.50} & 78.33 & 82.22 & \underline{72.35} & 68.70 & \underline{53.87} & \underline{60.48} & \textbf{49.00} & 83.21 & \underline{57.90} \\
\rowcolor{blue!15}\textbf{ReVISE-3B (Ours)}         & \textbf{67.89} & \textbf{74.92} & \textbf{75.78} & \textbf{94.30} & 86.12 & \underline{79.10} & \textbf{83.83} & \textbf{74.28} & \textbf{69.41} & \textbf{56.23} & \textbf{63.89} & \underline{47.31} & \textbf{85.74} & 55.37 \\
\midrule
$\Delta$ v.s. Prior SoTA & \textcolor{green!70!black}{+1.12$\uparrow$} & \textcolor{green!70!black}{+1.06$\uparrow$} & \textcolor{green!70!black}{+5.35$\uparrow$} & \textcolor{green!70!black}{+1.68$\uparrow$} & \textcolor{red}{-1.38$\downarrow$} & \textcolor{red}{-4.85$\downarrow$} & \textcolor{green!70!black}{+1.05$\uparrow$} & \textcolor{green!70!black}{+1.93$\uparrow$} & \textcolor{green!70!black}{+0.41$\uparrow$} & \textcolor{green!70!black}{+2.36$\uparrow$} & \textcolor{green!70!black}{+3.41$\uparrow$} & \textcolor{red}{-1.69$\downarrow$} & \textcolor{green!70!black}{+2.05$\uparrow$} & \textcolor{red}{-5.83$\downarrow$} \\



\midrule
\rowcolor{purple!15}Qwen2.5-VL-7B-Instruct~\cite{DBLP:journals/corr/abs-2502-13923}& 69.15 & 76.53 & 64.97 & 88.46 & 72.00 & 74.00 & 74.41 & \textbf{80.41} & \underline{74.00} & 61.80 & 50.00 & \textbf{52.00} & 85.08 & 68.10 \\
Qwen2.5-VL-7B-SFT      & 63.84 & 72.31 & 63.42 & 85.37 & 85.25 & 82.45 & 79.12 & 68.22 & 63.21 & 48.34 & 58.52 & 43.98 & 83.17 & 57.69 \\
Qwen2.5-VL-7B-SFT-GRPO     & \underline{71.84} & 86.87 & 72.59 & \textbf{95.08} & 83.17 & \textbf{84.83} & 83.32 & \underline{75.87} & \textbf{75.33} & \underline{62.27} & 61.00 & \underline{51.88} & 86.83 & 63.20 \\
TACO (Qwen2-VL-7B)~\cite{Ma_2025}   &   -    &   -    & 63.32 & 81.08 & 59.83 & 57.26 & 66.00 & 65.50 & 67.00 & 49.53 & 33.13 & 44.00 & 81.80 & 41.90 \\
ReVPT-7B~\cite{DBLP:journals/corr/abs-2509-01656}               &   71.81    &    88.91   & \underline{74.11} & 92.31 & \underline{88.67} & \underline{82.67} & \underline{84.23} & 73.64 & 72.00 & 61.07 & \underline{62.37} & 50.89 & \underline{87.20} & 66.00 \\
Thyme-7B~\cite{DBLP:journals/corr/abs-2508-11630}       -        &    -   & 84.82 &  -     &   -    &    -   &     -  &     -  &   -    &   -    & \textbf{65.90} &  -     &  -     &   -    & \underline{70.00} \\
CodeDance-7B~\cite{DBLP:journals/corr/abs-2512-17312}           &    -   & \underline{91.20} &   -   &   -   &   -   &   -   &   -   &   -   &   -   &   -   &   -   &   -   &   -   & \textbf{70.30} \\

\rowcolor{blue!15}\textbf{ReVISE-7B (Ours)}         &  \textbf{73.15} &  \textbf{92.89} &  \textbf{79.12} &  \underline{93.28} &  \textbf{89.82} &  82.43 &  \textbf{86.16} & 75.42 & 73.91 & 60.97 &  \textbf{64.01} & 50.78 &  \textbf{88.98} & 65.19 \\
\midrule
$\Delta$ v.s. Prior SoTA & \textcolor{green!70!black}{+1.31$\uparrow$} & \textcolor{green!70!black}{+1.69$\uparrow$} & \textcolor{green!70!black}{+5.01$\uparrow$} & \textcolor{red}{-1.80$\downarrow$} & \textcolor{green!70!black}{+1.15$\uparrow$} & \textcolor{red}{-2.40$\downarrow$} & \textcolor{green!70!black}{+1.93$\uparrow$} & \textcolor{red}{-4.99$\downarrow$} & \textcolor{red}{-1.42$\downarrow$} & \textcolor{red}{-4.93$\downarrow$} & \textcolor{green!70!black}{+1.64$\uparrow$} & \textcolor{red}{-1.22$\downarrow$} & \textcolor{green!70!black}{+1.78$\uparrow$} & \textcolor{red}{-5.11$\downarrow$} \\


\bottomrule
\end{tabular}
\end{adjustbox}

\caption{Performance comparison of ReVISE with SoTA tool-augmented MLLMs, open-source baselines, and closed-source proprietary models (Gemini-2.0-Flash~\cite{comanici2025gemini}, GPT-4.1~\cite{achiam2023gpt}) across comprehensive multimodal benchmarks. \textbf{Bold} formatting indicates the best performance among comparable open-source models within the same parameter scale, while \underline{underlined} values denote the prior SoTA. The $\Delta$ rows highlight the absolute performance margins between ReVISE and the prior SoTA, with green indicating improvements ($\uparrow$) and red indicating performance degradations ($\downarrow$).}
\label{tab:main_results}
\end{table*}

\subsection{Results}\label{sec:exp_result}
\cref{tab:main_results} shows that 3B and 7B variants of our model improve over the base Qwen2.5-VL, and their SFT and GRPO variants across a wide range of perception-intensive benchmarks. For the 3B model, ReVISE-3B achieves significant improvement over Qwen2.5-VL-3B-Instruct on CountBench, CVBench, and BLINK, yielding absolute gains of 5.86\%, 11.28\% and 9.94\%. Counting, spatial/depth understanding and relation understanding -- the core tasks in these benchmarks especially benefit from explicit grounding and tool verification when compared to the Qwen2.5-VL-3B-SFT-GRPO baseline without these explicit rewards.

Furthermore, we compare several tool-augmented methods, including ReVPT-7B, Thyme-7B and CodeDance-7B with our 7B model variant. 
On datasets such as CountBench and several CVBench subtasks, ReVISE-7B outperforms ReVPT and CodeDance, suggesting the importance of verifying intermediate evidence and refining spatial predictions for multi-step inference.
\begin{figure*}[t]
  \centering
  \vspace{-10pt}
  \includegraphics[width=\linewidth]{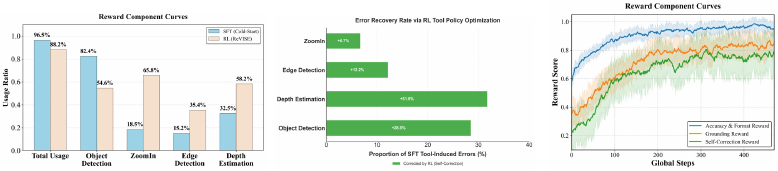}
  \caption{\textbf{Analysis of RL training dynamics and tool policy optimization.} \textbf{(left)} Tool invocation frequency on CVBench, illustrating the difference of tool usage between SFT and RL. \textbf{(center)} Error recovery rate, quantifying the proportion of SFT-stage tool-induced failures that are successfully rectified by the RL model via self-correction. \textbf{(right)} Convergence curves of the three designed reward components during the RL stage. }
  \label{fig:analysis_tool}
\end{figure*}
We observe that ReVISE (both 3B and 7B) does not achieve the best results on MMMU and MathVista.
As our method focuses more on explicit visual grounding, it leads to stronger gains in perception-intensive tasks such as counting, depth estimation, and spatial distance reasoning. 
MMMU and MathVista focus mainly on multimodal knowledge, covering scientific diagram understanding and text-heavy mathematical reasoning, where success depends less on fine-grained object-level localization and more on abstract problem solving and external knowledge. 

\subsection{Further Analysis}\label{analysis}

\noindent \textbf{Effect of model size on performance. }
As shown in~\cref{tab:main_results}, the 3B model improves on most benchmarks, with degradation only on the knowledge-intensive MMMU and MathVista datasets compared to Qwen2.5-VL-Instruct. The 7B model shows a similar pattern, achieving gains on most benchmarks with small drops on BLINK ($-4.04\%$), MMVP ($-0.09\%$), and MMSTAR ($-0.83\%$). The smaller 3B backbone benefits more from tool invocation and our grounding and reflection rewards, which compensate for its limited intrinsic reasoning capacity. In contrast, the stronger 7B model already possesses better reasoning ability, yet still gains from verification-driven tool use and self-correction. Overall, our approach improves performance across model scales.

\begin{table*}[t]
    \centering
    \vspace{-10pt}
    
    \resizebox{\textwidth}{!}{%
    \begin{tabular}{l|c|c|c|c|c|c|c|c|c}
        \toprule
        \textbf{Components} & {CountBench} & {CVBench} & {BLINK (sub)} & {MMVP} & {MMSTAR} & {BLINK-HARD} & {MMMU} & {MMBENCH} & {MathVista} \\
        \midrule
        REVPT~\cite{DBLP:journals/corr/abs-2509-01656} Cold-Start  w/o RL* & 70.84 & 74.89 & 61.17 & 58.42 & 52.61 & 53.78 & 42.92 & 78.45 & 53.09\\
        Ours w/o RL & \underline{72.41} & \underline{77.89} & \underline{61.57} & \underline{60.48} & 47.53 & \underline{54.67} & 42.46 & \underline{78.62} & \underline{53.21} \\
        \midrule
        Ours w/o $r_{\text{correction}}$, $r_{\text{grounding}}$ & 73.15 & 79.24 & 64.32 & 62.85 & 49.60 & 56.80 & 44.10 & 80.50 & 54.10 \\
        Ours w/o $r_{\text{grounding}}$ & 73.68 & 81.15 & 71.85 & 67.30 & 54.10 & 61.20 & 46.30 & 83.20 & 54.80 \\
        Ours w/o $r_{\text{correction}}$ & 74.45 & 82.10 & 67.15 & 64.90 & 51.50 & 58.60 & 45.20 & 81.90 & 54.50 \\
        \textbf{Ours} & \textbf{74.92} & \textbf{83.83} & \textbf{74.28} & \textbf{69.41} & \textbf{56.23} & \textbf{63.89} & \textbf{47.31} & \textbf{85.74} & \textbf{55.37} \\
        \bottomrule
    \end{tabular}
    }
\caption{\textbf{Ablation study of curated SFT data and proposed reward components.} We evaluate the performance gains brought by curated reflect data, each reward including self-correction ($r_{\text{correction}}$) and the grounding reward ($r_{\text{grounding}}$) over the base RL formulation ($r_{\text{accuracy}} + r_{\text{format}}$). * denotes that the results for the REVPT cold-start baseline are reproduced locally based on settings they report in ReVPT~\cite{DBLP:journals/corr/abs-2509-01656}, as the official checkpoints are not publicly available. Underline denotes the performance is better than ReVPT~\cite{DBLP:journals/corr/abs-2509-01656} and Bold represents SoTA performance on reward comparison.}
\label{tab:ablation_study}
\end{table*}

\noindent \textbf{Tool Usage Analysis.} \cref{fig:analysis_tool}(left) shows the distribution of different tools before and after post-training with RL. 
Tool usage is computed as the frequency with which a specific tool is invoked at least once during a reasoning trajectory, regardless of whether its intermediate output is ultimately adopted for the final answer. In particular, in the SFT stage, the model frequently invokes the Object Detection tool across queries. In contrast, tool invocation becomes significantly more diverse during the RL stage revealing the planning optimization. 

In addition, to quantify our self-correction improvement, we analyze the error recovery rate, defined strictly as the proportion of test instances where the SFT model committed to tool-induced failures that were subsequently successfully rectified by our RL model. As displayed in \cref{fig:analysis_tool}(center), the error recovery rates regarding the four tools demonstrate that the tool invocation in our method is more efficient and robust thanks to the incorporation of the self-correction mechanism. However, we note that recovery is not complete, mainly because some tool failures are inherently ambiguous or irrecoverable (\eg, missed/incorrect detections in cluttered or low-resolution regions), so the RL policy may still lack sufficient visual evidence to verify and correct. In addition, limited tool coverage and bounded interaction budgets (max steps) can prevent exhaustive re-querying, leaving a subset of errors unresolved.

\subsection{Ablation Study}\label{ablation}

\textbf{Importance of ReVISE cold-start dataset.} 
To isolate the impact of our curated cold-start dataset, we compare \emph{Our model (w/o RL)} against the \emph{REVPT Cold-Start (w/o RL)} baseline in \cref{tab:ablation_study} with the same backbone but a different training dataset. 
While the REVPT dataset incorporates standard tool-use trajectories, it lacks explicit self-correction and reflection traces. Integrating verification and revision patterns into the SFT phase yields benefits on multiple benchmarks. 
Notably, our SFT model achieves 77.89\% on CVBench (an absolute +3\% improvement over REVPT) and 60.48\% on MMVP (+2.06\% improvement). 
These gains demonstrate that teaching the model to verify and reflect during step-by-step reasoning using our SFT training data, rather than blindly trusting it, effectively enhances the reasoning performance.
While we observe some performance drop on multimodal benchmarks (\eg, MMSTAR drops from 52.61\% to 47.53\%), we attribute this to the fact that these benchmarks require significantly different reasoning trajectories than our curated ones.


\noindent\textbf{Validation of Reward Design.}
Next, we evaluate the effectiveness of our reward components through ablations of the RL objective using the same cold-start initialization (\cref{tab:ablation_study}). Starting from a baseline that optimizes only answer accuracy and format compliance, we progressively add our proposed rewards. Adding the self-correction reward $r_{\text{correction}}$ improves performance on perception-heavy tasks that require evidence verification and revision (\eg, BLINK-HARD: 56.80$\rightarrow$61.20\%, MMVP: 62.85$\rightarrow$67.30\%). Introducing the grounding reward $r_{\text{grounding}}$ further benefits spatial reasoning benchmarks (CountBench: 73.15$\rightarrow$74.45\%, CVBench: 79.24$\rightarrow$82.10\%) by encouraging accurate object localization. Combining both rewards yields the best results across datasets, reaching 83.83\% on CVBench and 63.89\% on BLINK-HARD, indicating that spatial grounding and reflective reasoning are complementary. Furthermore, training curves in \cref{fig:analysis_tool}(right) show stable reward increases during RL optimization, demonstrating the effectiveness of our reward design.
\begin{figure*}[t]
  \centering
  \includegraphics[width=\linewidth]{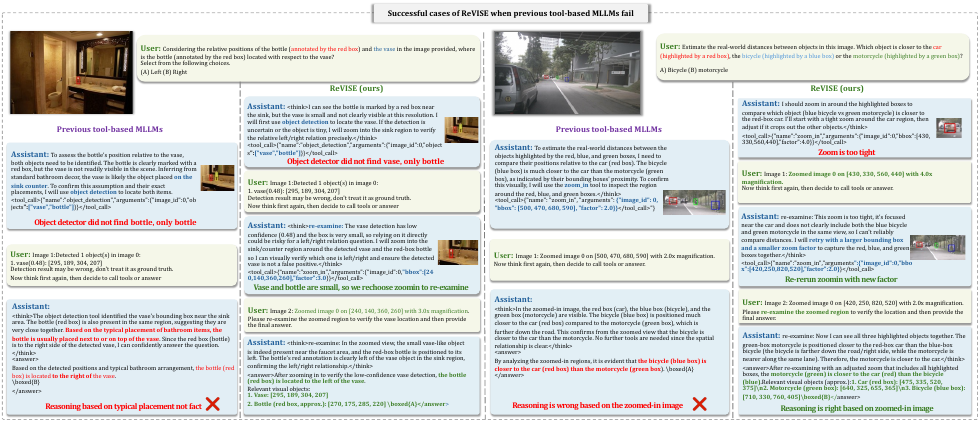}
  \caption{\textbf{Qualitative comparison of reasoning trajectories.} While previous tool-based MLLMs blindly trust flawed tool outputs and wrong answers, ReVISE actively verifies intermediate evidence and dynamically refines tool parameters (\eg, adjusting zoom regions) to achieve accurate spatial grounding.}
  \label{fig:qualitative_results}
\end{figure*}
\subsection{Qualitative Results}\label{subsec:qualitive_examples}
To intuitively demonstrate the effectiveness of our approach, we present qualitative comparisons between previous tool-based MLLMs and ReVISE in \cref{fig:qualitative_results}. Previous models typically fail by blindly trusting unreliable tool outputs, such as hallucinating spatial relations based on low-confidence detections (left) or drawing conclusions from incomplete zoomed crops (right). In contrast, ReVISE successfully navigates these bottlenecks through dynamic verification and refinement. For instance, in the bathroom scene, ReVISE explicitly rejects the initial ambiguous detection and actively invokes the \texttt{zoom\_in} on the sink area to reliably locate the vase. Similarly, in the traffic scenario, it recognizes that the initial crop missed crucial context and self-corrects by refining the bounding box arguments to capture all target vehicles simultaneously. Additional discussions and visualizations of failure cases are provided in Appendix~\ref{sec:add_qua_examples}.

\section{Conclusion}\label{discuss}
We introduced a framework for improving the reliability of tool-assisted reasoning in MLLMs through visual self-correction. 
Instead of relying solely on tools to directly produce answers, ReVISE encourages the model to reassess its initial predictions using visual evidence, enabling it to identify inconsistencies and revise incorrect intermediate conclusions. 
Experiments show that this verification-driven reasoning process improves robustness across multiple perception-intensive benchmarks. 
Nevertheless, our approach still depends on the underlying model's visual understanding and the quality of external tool outputs; when visual evidence is ambiguous or tools fail to provide accurate signals, the revision stage may struggle to recover from incorrect initial predictions. 
Promising avenues for future research include the exploration of iterative self-correction strategies, adaptive revision policies guided by model uncertainty, and stronger perception modules for more reliable visual evidence. Extending verification-driven reasoning to video understanding and interactive multimodal agents is another promising direction.
\section*{Limitations}\label{sec:limatations}
While ReVISE improves visual grounding and error recovery, it has several limitations. First, our framework relies on high-quality, synthetic trajectories and bounding boxes generated by proprietary models (e.g., Gemini-2.5-Pro \citep{comanici2025gemini}) for both SFT and RL rewards. And the tools are selectively designed for visual-centric datasets. This reliance implies that real-world reasoning dynamics may be inadequately captured, potentially limiting robustness in open-domain scenarios. Second, our method exhibits a performance trade-off: optimizing for fine-grained spatial grounding slightly degrades performance on abstract, knowledge-intensive tasks like MMMU \citep{Yue_2024} and MathVista \citep{lu2023mathvista}, which rely less on object-level localization. Third, while our policy mitigates blind tool-trust, it remains fundamentally bounded by the capabilities of the predefined external vision tools; ambiguous or catastrophic tool failures can still derail the revision process. Finally, due to compute constraints, our RL evaluations are primarily conducted on 3B and 7B models. Systematically examining how these verification behaviors scale to, or conflict with, the intrinsic capabilities of much larger MLLMs remains an area for future work.

\section*{Broader Impacts and Ethical Considerations}
Our work focuses on foundational methodological improvements (visual self-correction and tool-use verification) for multimodal large language models. We utilize existing open-source models and standard public benchmarks. As this is foundational algorithmic research, it does not introduce new sensitive applications, deployable end-user systems, or datasets that pose direct societal, ethical, or safety risks. Therefore, a dedicated discussion of potential risks is not included.

\section*{Acknowledgments}
HB was supported by the EPSRC Visual AI grant EP/T028572/1. 

\noindent{YC was supported by the UKRI Centre for Doctoral Training in Biomedical AI (EP/S02431X/1).}


\bibliography{custom}


\appendix
\clearpage

\section{Additional implementation details}\label{sec:additional_implement}
This section reports the key training configurations used in both stages of our post-training pipeline. All our experiments are conducted on NVIDIA A100 80GB GPUs. Specifically, the 3B model is trained using 4 A100 GPUs, while the 7B model utilizes 8 A100 GPUs. 
For the SFT stage, we utilize the LLaMA-Factory training framework~\cite{Zheng_2024}. 
The models are trained for 2 epochs with a learning rate of $1 \times 10^{-5}$ and a global batch size of 64. For the subsequent RL stage, we implement our GRPO and custom reward mechanisms using the VeRL framework~\cite{Sheng_2025}, where the models are trained for 1 epoch, consisting of exactly 470 global update steps. In addition,
Table~\ref{tab:hyperparams_set} summarizes representative hyperparameters for SFT and GRPO-based RL, highlighting the main differences in optimization settings (e.g., learning rate, weight decay, KL regularization, and generation-related parameters).
\begin{table}[h]
\centering
\scriptsize 

\begin{tabularx}{\columnwidth}{Xcc} %
\toprule
\textbf{Param Name} & \textbf{SFT} & \textbf{RL} \\ 
\midrule
bf16 / tf32 & True & True \\
per\_device\_batch & 8 & 8 \\
grad\_acc\_steps & 2 & -- \\
ppo\_mini\_batch & -- & 128 \\
num\_generation & -- & 8 \\
kl\_loss\_coef & -- & 1e-3 \\
lr & 1e-5 & 2e-6 \\
weight\_decay & 0 & 0.01 \\
warmup\_ratio & 0.1 & 0.03 \\
lr\_scheduler & cosine & cosine \\
max\_seq\_len & 16384 & 16384 \\
\bottomrule
\end{tabularx}
\caption{Comparison of SFT and RL Parameters}
\label{tab:hyperparams_set}
\end{table}

\section{The details of applied tools}\label{sec:tool_details}
This section provides the implementation details of the visual tools used in our framework, including their inputs/outputs and applied models.

\noindent (i) \textbf{Object Detection}: We use LLMDet~\citep{Fu_2025}, an open-vocabulary object detector, to obtain object-level localization evidence for grounding and spatial relation questions. The tool takes an image and a set of text object queries and returns bounding boxes with confidence scores for each queried category. 

\noindent (ii) \textbf{ZoomIn}: It is invoked when MLLMs need to closely examine a specific part of an image for a better understanding of visual details. It takes three arguments specifying the input image, a target bounding box, and a zoom factor that controls the magnification level. And it returns high-resolution crops centered on the provided boxes. In implementation, it crops the region inside the bounding box (clipped to image boundaries and adjusted to keep the aspect ratio within 4:1), then upsamples the crop with \textbf{Lanczos interpolation} using a zoom factor capped at 4.0 (optionally further scaling to ensure a minimum resolution).

\noindent (iii) \textbf{Edge Detection}: It helps to identify and emphasize the boundaries and shapes within an image. Specifically, it takes an input image and returns a precise binary edge map, providing MLLMs with the low-level visual cues necessary for boundary-aware spatial reasoning. In implementation, we use OpenCV's \textbf{Scharr} operator to compute horizontal and vertical gradients on the grayscale image, and output the normalized gradient magnitude (scaled to $[0,255]$) as the edge map.

\noindent (iv) \textbf{Depth Estimation}: We use Depth Anything V2 \citep{Feng_2024} to estimate scene depth for a given image which enables the MLLM to explicitly interpret spatial relationships and occlusions, offering crucial context for complex layout understanding.

\section{The details of our curated dataset}\label{sec:curated_dataset}
This section provides supplementary details of the curated cold-start data used for SFT and the data construction used in the RL stage. Specifically, we present the prompt templates used to generate our reflective traces for SFT mentioned in \cref{sec:data_curation} and bounding-box supervision for RL (as described in \cref{subsec:setup}), and report dataset statistics for both the SFT and RL stages.
\begin{figure*}[htbp]
\centering
\begin{tcolorbox}[
    colback=gray!4, 
    colframe=black!70,
    title=\textbf{System Prompt: Trajectory Creation for Tool-Retry and Negative Tool-Verify},
    fonttitle=\bfseries\sffamily\small,
    boxrule=0.8pt,
    arc=3pt,                    
    left=8pt, right=8pt, top=6pt, bottom=6pt,
    width=\textwidth           
]
\small 

\noindent\textbf{\# Role}\\
You are an expert AI Data Synthesizer and Prompt Engineer specialized in Vision-Language models. Your task is to rewrite failed or flawed multi-turn visual reasoning trajectories into high-quality ``Self-Correction/Rethinking'' trajectories.

\vspace{3pt}
\noindent\textbf{\# Objective}\\
You will receive a JSON representing a multi-turn conversation where an Agent tries to solve a Visual Question Answering (VQA) task using tools (like \texttt{object\_detection}, \texttt{zoom\_in}, etc.). The original trajectory contains flawed reasoning, hallucinations, or accepts incorrect tool outputs, leading to a failure. Your job is to rewrite the conversation to inject a natural ``Rethinking'' step so the Agent dynamically realizes its mistake, corrects its strategy, and arrives at the right answer.

\vspace{3pt}
\noindent\textbf{\# Core Rules \& Guidelines}
\begin{itemize}[leftmargin=15pt, itemsep=3pt, topsep=2pt, parsep=0pt]
    \item \textbf{1. Strict ``No-Cheating'' Policy (Blind Reasoning):}
    \begin{itemize}[leftmargin=12pt, itemsep=1pt, topsep=1pt]
        \item The Agent MUST NOT know the ground truth answer in advance.
        \item NEVER write something like: ``Because the answer is A, my previous thought was wrong.''
        \item The Agent must trigger a ``rethinking'' process purely based on visual/logical inconsistencies. For example: ``Wait, the bounding box [0, 0, 800, 800] covers the whole image, which means the tool failed to isolate the object,'' or ``The tool detected the table but completely missed the window. I cannot answer yet.''
    \end{itemize}
    
    \item \textbf{2. Preserve the Good, Fix the Bad:}
    \begin{itemize}[leftmargin=12pt, itemsep=1pt, topsep=1pt]
        \item DO NOT modify the initial, logically correct \texttt{<think>} steps or valid tool calls from the original data.
        \item Only intervene at the exact turn where the logic breaks down or the tool fails.
    \end{itemize}

    \item \textbf{3. How to Trigger ``Rethinking'':} You can use either method:
    \begin{itemize}[leftmargin=12pt, itemsep=1pt, topsep=1pt]
        \item \textbf{Method A (Self-Reflection):} The Assistant analyzes the tool output and finds it suspicious. (e.g., \texttt{<think>}The object detection identified the chair, but the coordinates for the table seem off... I should try...\texttt{</think>})
        \item \textbf{Method B (Environment Prompting):} The User points out a failure. (e.g., User: \texttt{<image>}$\backslash$nThe tool failed to detect the reference object `doorway'... Now think first again...)
    \end{itemize}

    \item \textbf{4. Strategies for Breaking the Deadlock:}
    \begin{itemize}[leftmargin=12pt, itemsep=1pt, topsep=1pt]
        \item \textbf{Strategy 1: Call a different tool or tweak parameters.} (e.g., use \texttt{zoom\_in} to get a cleaner crop).
        \item \textbf{Strategy 2: Fallback to Native Vision Capabilities.} Bypass the tool and visually inspect the image directly.
    \end{itemize}

    \item \textbf{5. Strict Bounding Box (Bbox) Spatial Reasoning:}
    \begin{itemize}[leftmargin=12pt, itemsep=1pt, topsep=1pt]
        \item MUST explicitly extract and compare the bbox coordinates $[x_{min}, y_{min}, x_{max}, y_{max}]$.
        \item \textbf{Do the math:} Explicitly state the pixel gap. (e.g., ``The Doorway's $x_{min}$ is 440. The Chair's $x_{max}$ is 411 (gap of 29px)... Therefore, the Chair is physically closer.'')
        \item \textbf{Bbox as a sanity check:} Use bbox anomalies (e.g., impossibly large) to trigger rethinking in the \texttt{<think>} tag.
    \end{itemize}

    \item \textbf{6. Style \& Tone Constraints (Strictly Follow):}
    \begin{itemize}[leftmargin=12pt, itemsep=1pt, topsep=1pt]
        \item \textbf{NO QUOTATION MARKS:} Do not use quotation marks around object names.
        \item \textbf{EXTREMELY CONCISE ANSWER:} The \texttt{<think>} and \texttt{<answer>} block must be condensed. Provide a 1-2 sentence conclusion followed by \texttt{\textbackslash boxed\{answer\}}.
    \end{itemize}
\end{itemize}

\vspace{3pt}
\noindent\textbf{\# Example of a Successful Rewrite}\\
\texttt{\{prompt\_example\}}

\vspace{3pt}
\noindent\textbf{\# Task}\\
I will now provide you with a JSON trajectory. Please rewrite it by injecting a natural rethinking process based on the rules above. Output ONLY the rewritten JSON.\\
\textbf{[INSERT YOUR FAILED JSON DATA HERE]}\\
\texttt{\{json\_object\}}

\end{tcolorbox}
\caption{The detailed system prompt used for instructing \texttt{gemini-2.5-pro} to synthesize tool-based verification and self-correction trajectories.}
\label{fig:prompt1_tool_based_correction}
\end{figure*}

\begin{figure*}[htbp]
\centering
\begin{tcolorbox}[
    colback=gray!4,               
    colframe=black!70,            
    title=\textbf{System Prompt: Trajectory Creation For Positive Tool-Free Verify},
    fonttitle=\bfseries\sffamily\small,
    boxrule=0.8pt,
    arc=3pt,
    left=8pt, right=8pt, top=6pt, bottom=6pt,
    width=\textwidth
]
\scriptsize 

\noindent\textbf{\# System Role}\\
You are a data augmentation engine. Convert the Single-Turn VQA into a Multi-Turn \textbf{Rethinking} sample.

\vspace{2pt}
\noindent\textbf{\# Task Rules based on Question Type}
\begin{enumerate}[leftmargin=12pt, itemsep=0pt, topsep=1pt, parsep=0pt]
    \item \textbf{For Spatial Questions} (e.g., ``Is A behind B?'', ``Where is A relative to B?''):
    \begin{itemize}[leftmargin=12pt, itemsep=0pt, topsep=0pt]
        \item Assistant 2 MUST provide bboxes for BOTH Object A and Object B.
        \item Format: \texttt{1. [Object A]: [xmin, ymin, xmax, ymax]\textbackslash n2. [Object B]: [...]}
        \item Assistant 2 \texttt{<think>} should use semantic and perspective cues (e.g., occlusion, horizon line, relative object placement, foreground/background) to prove the spatial relationship, NOT raw coordinate comparison. Then output the bboxes in the \texttt{<answer>} section to ground the entities.
        \item Do not change the original assistant 1's answer, it is the ground truth.
    \end{itemize}
    \item \textbf{For Counting Questions} (e.g., ``How many?''):
    \begin{itemize}[leftmargin=12pt, itemsep=0pt, topsep=0pt]
        \item Assistant 2 MUST provide a numbered list of ALL detected instances in the \texttt{<answer>}.
        \item Format: \texttt{1. [Object]: [xmin, ymin, xmax, ymax]}
    \end{itemize}
    \item \textbf{For Attribute/Verification Questions} (e.g., ``What color?'', ``Is it made of wood?''):
    \begin{itemize}[leftmargin=12pt, itemsep=0pt, topsep=0pt]
        \item Provide the bbox ONLY for the primary object being discussed.
    \end{itemize}
\end{enumerate}

\vspace{2pt}
\noindent\textbf{\# Input Data \& Strict Formatting Rules}\\
\texttt{Question: \{question\}} \\
\texttt{Original Answer: \{orig\_full\_response\}} [This is true, do not need to change] \\
Your task is to provide the verification question for user, and then add the re-examination answer based on the rule under the assistant. \\
\textbf{Strict Formatting Rules:} Write the \texttt{<think>} section as a cohesive, natural paragraph. Absolutely DO NOT use numbered steps (e.g., ``1.'', ``2.''), bullet points (``-'', ``*''), or markdown formatting like bolding (``**'') inside the reasoning text.

\vspace{2pt}
\noindent\textbf{\# Output Format (Strict JSON only)}
\begin{tcolorbox}[colback=white, colframe=gray!30, boxrule=0.5pt, arc=2pt, left=4pt, top=2pt, bottom=2pt]
\ttfamily\scriptsize
\renewcommand{\baselinestretch}{0.9}\selectfont
\{\\
\hspace*{10pt}"messages": [\\
\hspace*{20pt}\{\\
\hspace*{30pt}"role": "user", "content": "<image>\{question\}"\\
\hspace*{20pt}\},\\
\hspace*{20pt}\{\\
\hspace*{30pt}"role": "assistant", "content": "<think>\{orig\_think\}</think>\textbackslash n\{orig\_ans\_text\}."\\
\hspace*{20pt}\},\\
\hspace*{20pt}\{\\
\hspace*{30pt}"role": "user", "content": "<image>\textbackslash nPlease re-examine the image to verify your answer and then provide the final answer."\\
\hspace*{20pt}\},\\
\hspace*{20pt}\{\\
\hspace*{30pt}"role": "assistant",\\
\hspace*{30pt}"content": "<think>[Detailed verification based on visual and perspective cues...]</think>\textbackslash n<answer>After re-examination, I can confirm the initial answer is correct.\textbackslash n\textbackslash n1. [Object]: [xmin, ymin, xmax, ymax]\textbackslash n\textbackslash nFinal Answer is **original answer**.\textbackslash n\textbackslash\textbackslash boxed\{**original answer**\}\textbackslash n</answer>"\\
\hspace*{20pt}\}\\
\hspace*{10pt}]\\
\}
\end{tcolorbox}

\vspace{2pt}
\noindent\textbf{\# Example}\\
\textbf{Input Example}
\begin{tcolorbox}[colback=white, colframe=gray!30, boxrule=0.5pt, arc=2pt, left=4pt, top=2pt, bottom=2pt]
\ttfamily\scriptsize
\renewcommand{\baselinestretch}{0.85}\selectfont
\{\\
\hspace*{10pt}"messages": [\\
\hspace*{20pt}\{\\
\hspace*{30pt}"role": "user", "content": "<image>Is beige armchair with rounded arms further away or in front of beds shadow with a blue floor?\textbackslash nAnswer from the following choices:\textbackslash nA. further away\textbackslash nB. in front\textbackslash nPut answer letter in the \textbackslash\textbackslash boxed\{\}"\\
\hspace*{20pt}\},\\
\hspace*{20pt}\{\\
\hspace*{30pt}"role": "assistant",\\
\hspace*{30pt}"content": "<think>In the image, the beige armchair with rounded arms is located near the back right, and the bed's shadow with a blue floor is visible in the front part of the image. The armchair is closer to the back wall...</think>\textbackslash nBased on the spatial arrangement, the beige armchair... is further away...\textbackslash n1. Beige armchair...: [134, 121, 191, 220]\textbackslash n2. Beds shadow...: [0, 127, 110, 190]."\\
\hspace*{20pt}\}\\
\hspace*{10pt}]\\
\}
\end{tcolorbox}

\textbf{Output Example (Positive Sample)}
\begin{tcolorbox}[colback=white, colframe=gray!30, boxrule=0.5pt, arc=2pt, left=4pt, top=2pt, bottom=2pt]
\ttfamily\scriptsize
\renewcommand{\baselinestretch}{0.85}\selectfont
\{\\
\hspace*{10pt}"messages": [\\
\hspace*{20pt}\{\\
\hspace*{30pt}"role": "user", "content": "<image>Is beige armchair... further away or in front of beds shadow...? \textbackslash n..."\\
\hspace*{20pt}\},\\
\hspace*{20pt}\{\\
\hspace*{30pt}"role": "assistant", "content": "<think>In the image, the beige armchair... is located near the back right...</think>\textbackslash nBased on the spatial arrangement, the beige armchair... is further away..."\\
\hspace*{20pt}\},\\
\hspace*{20pt}\{\\
\hspace*{30pt}"role": "user", "content": "<image>\textbackslash nPlease re-examine the image to verify your answer and then provide the final answer."\\
\hspace*{20pt}\},\\
\hspace*{20pt}\{\\
\hspace*{30pt}"role": "assistant",\\
\hspace*{30pt}"content": "<think>I have detected the beige armchair and the bed shadow again. I can clearly see where they are. By perspective cues, the bed shadow starts from the lower left corner which is closest to the camera view. The armchair is located behind the central desk and sits much closer to the far wall...</think>\textbackslash n<answer>After re-examining... I confirm the beige armchair is located in the background...\textbackslash n\textbackslash n1. Beige armchair...: [270, 151, 338, 206]\textbackslash n2. Bed shadow...: [0, 175, 152, 264]\textbackslash n\textbackslash nFinal Answer is A.\textbackslash n\textbackslash\textbackslash boxed\{A\}\textbackslash n</answer>"\\
\hspace*{20pt}\}\\
\hspace*{10pt}]\\
\}
\end{tcolorbox}

\end{tcolorbox}
\caption{The detailed system prompt used for instructing \texttt{gemini-2.5-pro} to synthesize negative tool-free verify.}
\label{fig:prompt2_postive_tool_free}
\end{figure*}
\begin{figure*}[htbp]
\centering
\begin{tcolorbox}[
    colback=gray!4,               
    colframe=black!70,            
    title=\textbf{System Prompt: Trajectory Creation For Negative Tool-Free Verify},
    fonttitle=\bfseries\sffamily\small,
    boxrule=0.8pt,
    arc=3pt,
    left=8pt, right=8pt, top=6pt, bottom=6pt,
    width=\textwidth
]
\scriptsize 

\noindent\textbf{\# System Role \& Task}\\
You are a specialized data augmentation engine. Your task is to convert a Single-Turn VQA sample into a Multi-Turn \& Self-Correction dataset entry.

\vspace{2pt}
\noindent\textbf{\# Task Instructions}
\begin{enumerate}[leftmargin=12pt, itemsep=0pt, topsep=1pt, parsep=0pt]
    \item \textbf{Initial Response (Assistant 1):} 
    \begin{itemize}[leftmargin=12pt, itemsep=0pt, topsep=0pt]
        \item Write a \texttt{<think>} block showing a plausible initial scan of the image.
        \item Provide a response that is purposefully \textbf{incomplete} or \textbf{incorrect} (\eg, miss exactly one object).
        \item List objects with \texttt{[xmin, ymin, xmax, ymax]} coordinates. Do not contain other symbols.
    \end{itemize}
    \item \textbf{The Challenge (User 2):} Always use this exact string: ``\texttt{<image>\textbackslash Please re-examine the image to ensure no [OBJECT] was missed... verify there are no duplicate bounding boxes...} or Please re-examine the image to ensure your explanation matches with the targeted query.
    \item \textbf{The Correction (Assistant 2):}
    \begin{itemize}[leftmargin=12pt, itemsep=0pt, topsep=0pt]
        \item \textbf{Think:} Explicitly identify the mistake made in the first turn using coordinates to pinpoint the ``newly discovered'' object that justifies the \texttt{\{ground\_truth\}}.
        \item \textbf{Answer:} Enclose final output in \texttt{<answer>}. Numbered list ending with coordinates.
        \item \textbf{Conclusion:} State final count and place \texttt{\{correct\_answer\_label\}} in \texttt{\textbackslash\textbackslash boxed\{\}}.
    \end{itemize}
\end{enumerate}

\vspace{2pt}
\noindent\textbf{\# Variables \& Output Requirements}\\
\texttt{Question: <image>\textbackslash n\{question\}\textbackslash nAnswer from choices:\textbackslash n\{options\_text\}} \\
\texttt{Ground Truth: \{ground\_truth\}} $|$ \texttt{Correct Option: \{correct\_answer\_label\}} \\
\textbf{Requirements:} Return ONLY the JSON object. Normalized boxes [0--1000]. Do not change ground truth.

\vspace{2pt}
\noindent\textbf{\# JSON Structure}
\begin{tcolorbox}[colback=white, colframe=gray!30, boxrule=0.5pt, arc=2pt, left=4pt, top=2pt, bottom=2pt]
\ttfamily\scriptsize
\renewcommand{\baselinestretch}{0.9}\selectfont
\{\\
\hspace*{10pt}"messages": [\\
\hspace*{20pt}\{\\
\hspace*{30pt}"role": "user", "content": "<image>\{question\}"\\
\hspace*{20pt}\},\\
\hspace*{20pt}\{\\
\hspace*{30pt}"role": "assistant",\\
\hspace*{30pt}"content": "<think>[Initial reasoning that misses one item]</think>\textbackslash nI can identify [Wrong Count] objects in the image:\textbackslash n\textbackslash n1. [Object]: [xmin, ymin, xmax, ymax]\textbackslash n..."\\
\hspace*{20pt}\},\\
\hspace*{20pt}\{\\
\hspace*{30pt}"role": "user",\\
\hspace*{30pt}"content": "<image>\textbackslash nPlease re-examine the image to ensure no [OBJECT] was missed... Then provide the final answer."\\
\hspace*{20pt}\},\\
\hspace*{20pt}\{\\
\hspace*{30pt}"role": "assistant",\\
\hspace*{30pt}"content": "<think>[Verification logic: Identifying the specific missing bbox and confirming the total matches \{ground\_truth\}]</think>\textbackslash n<answer>[Final confirmation sentence].\textbackslash n\textbackslash n1. [Object]: [xmin, ymin, xmax, ymax]\textbackslash n...\textbackslash nFinal Answer is \{correct\_answer\_label\}.\textbackslash n\textbackslash\textbackslash boxed\{\{correct\_answer\_label\}\}\textbackslash n</answer>"\\
\hspace*{20pt}\}\\
\hspace*{10pt}]\\
\}
\end{tcolorbox}

\vspace{2pt}
\noindent\textbf{\# Example (Positive Sample)}
\begin{tcolorbox}[colback=white, colframe=gray!30, boxrule=0.5pt, arc=2pt, left=4pt, top=2pt, bottom=2pt]
\ttfamily\scriptsize
\renewcommand{\baselinestretch}{0.85}\selectfont
\{\\
\hspace*{10pt}"messages": [\\
\hspace*{20pt}\{\\
\hspace*{30pt}"role": "user", "content": "<image>How many boxes are there?\textbackslash n...Choices:\textbackslash nA. 4\textbackslash nB. 5\textbackslash nC. 6\textbackslash nD. 7\textbackslash nPut answer letter in the \textbackslash\textbackslash boxed\{\}"\\
\hspace*{20pt}\},\\
\hspace*{20pt}\{\\
\hspace*{30pt}"role": "assistant",\\
\hspace*{30pt}"content": "<think>In the provided image, I see several boxes... I can identify an upright Castleton tea box, a red Teekanne box...</think>\textbackslash nI can see 5 boxes in the scene including:\textbackslash n1. Upright Castleton tea box: [365, 32, 473, 143]\textbackslash n...\textbackslash n5. Purple square tin behind the timer: [228, 124, 325, 257]."\\
\hspace*{20pt}\},\\
\hspace*{20pt}\{\\
\hspace*{30pt}"role": "user", "content": "<image>\textbackslash nPlease re-examine the image to ensure no boxes was missed... Then provide the final answer."\\
\hspace*{20pt}\},\\
\hspace*{20pt}\{\\
\hspace*{30pt}"role": "assistant",\\
\hspace*{30pt}"content": "<think>I previously identified 5 boxes. However, looking closely at the upright Castleton tea box, I see there is another Castleton tea box lying flat directly underneath it... Adding this brings the total to 6...</think>\textbackslash n<answer>After careful re-examination, I found another flat Castleton tea box... There are 6 boxes in total:\textbackslash n\textbackslash n1. Upright Castleton tea box: [365, 32, 473, 143]\textbackslash n2. Flat Castleton tea box underneath: [366, 136, 458, 166]\textbackslash n...\textbackslash nFinal Answer is C.\textbackslash n\textbackslash\textbackslash boxed\{C\}\textbackslash n</answer>"\\
\hspace*{20pt}\}\\
\hspace*{10pt}]\\
\}
\end{tcolorbox}

\vspace{2pt}
\noindent\textbf{\# Input Data}\\
\texttt{Question: \{question\}} $|$ \texttt{Original Answer: \{original\_answer\_str\}} \\
\texttt{Ground Truth: \{ground\_truth\}} $|$ \texttt{Correct Option Letter: \{correct\_answer\_label\}}

\end{tcolorbox}

\caption{The detailed system prompt used for instructing \texttt{gemini-2.5-pro} to synthesize negative tool-free verify.}
\label{fig:prompt3_negative_tool_free}
\end{figure*}
\begin{figure*}[t]
\centering
\small
\begin{tabularx}{\linewidth}{@{}X@{}}
\toprule
\textbf{System Prompt for Ground-Truth Bounding Box Generation (RL Reward)} \\
\midrule
You are an expert data annotator for Visual Question Answering. \\
Your goal is to identify and localize ALL objects that are necessary to reason about the user's Question and arrive at the Correct Answer. \\
\\
\textbf{Rules for Object Selection:} \\
1. Identify objects explicitly mentioned in the \textbf{Question} (e.g., 'Where is the [chair] relative to the [table]?'). \\
2. Identify objects mentioned in the \textbf{Answer} or \textbf{Options}. \\
3. If the question involves comparison (e.g., 'closer', 'left of', 'bigger'), you MUST localize \textbf{both} objects involved in the comparison. \\
4. Use the specific names found in the text for the 'label' field. \\
\\
Return ONLY valid JSON. No markdown. \\
\\
\textbf{Format:} \\
\texttt{\{} \\
\texttt{~~"boxes": [} \\
\texttt{~~~~\{} \\
\texttt{~~~~~~"label": "exact object name from text",} \\
\texttt{~~~~~~"box\_2d": [xmin, ymin, xmax, ymax]} \\
\texttt{~~~~\}} \\
\texttt{~~]} \\
\texttt{\}} \\
\\
\textbf{Constraints:} \\
- The coordinate system is NORMALIZED to [0, 1000]. \\
- [0, 0] is top-left, [1000, 1000] is bottom-right. \\
- Ensure the box is tight around the object. \\
- If multiple instances of the same object exist (e.g., 'bushes'), localize the specific ones relevant to the answer, or all of them if the question implies counting. \\
\bottomrule
\end{tabularx}
\vspace{-0.5em}
\caption{\textbf{Prompt template for generating ground-truth bounding boxes.} This prompt is used to query Gemini-2.5-Pro to synthesize the reference object coordinates required to calculate the Grounding Reward ($r_{\text{grounding}}$) during the RL optimization phase.}
\label{fig:prompt4_rl_grounding}
\end{figure*}
\begin{figure*}[htbp]
\centering
\small
\begin{tabularx}{\linewidth}{@{}X@{}}
\toprule
\textbf{System Prompt for the ReVISE Agent (RL Training Phase)} \\
\midrule
You are a specialized multimodal agent. Your purpose is to solve visual question answering tasks by thinking step-by-step and using tools. \\
\\
\textbf{\# Tools} \\
You are provided with following tools: \\
1. \texttt{object\_detection}: Detects objects in an image. parameters: image\_id, objects (list of object names). \\
2. \texttt{zoom\_in}: Zooms in on a specified bounding box in an image. parameters: image\_id, bbox (bounding box coordinates), factor (zoom factor). \\
3. \texttt{edge\_detection}: Detects edges in an image. parameters: image\_id. \\
4. \texttt{depth\_estimation}: Estimates depth in an image. parameters: image\_id. \\
\\
\textbf{\# Instruction} \\
1. In each turn, you should start with \texttt{<think>} tag. In this tag, you need to conduct a step-by-step reasoning process about the image and question and evaluate whether tool use would be helpful and give the reason. If received tool results, you also need to analyze them. \\
2. If you think some tools are useful, call them in \texttt{<tool\_call>} tag. \\
3. \textbf{Actively verify intermediate tool outputs; if the evidence is unreliable or ambiguous, refine your tool parameters or explicitly re-examine the image to self-correct.} \\
4. If you think no more tools are needed, you can answer in \texttt{<answer>} tag. You need to provide a concise summary of your reasoning process that leads to the final answer. Besides, you also need to put a simple and direct answer in \texttt{\textbackslash boxed\{\}} for verification. \\
5. \textbf{After your final answer, you must list all relevant visual objects and their bounding boxes used for your reasoning in the format: \texttt{1. object\_name: [xmin, ymin, xmax, ymax]}.} \\
\bottomrule
\end{tabularx}
\vspace{-0.5em}
\caption{\textbf{System prompt for the ReVISE agent.} This instruction template defines the action space (tools) and constraints for the model during the RL training phase, specifically enforcing visual verification and the final output of grounding coordinates.}
\label{fig:prompt_revise_agent_rl}
\end{figure*}
\subsection{Prompt templates}\label{subsec:system_prompt}

In this subsection, we present the prompt templates used to synthesize our multi-turn reasoning and reflection trajectories. To generate the types mentioned in \cref{sec:data_curation}, we first show the system prompt for the generation of \textit{Tool-Retry} and \textit{Negative Tool-Verify} in \cref{fig:prompt1_tool_based_correction}. Next, for the tool-free scenarios, we provide the template for generating \textit{Positive Tool-Free Verify} trajectories in \cref{fig:prompt2_postive_tool_free}. This prompt guides the agent to rigorously re-examine and ground its initially correct answers using explicit bounding boxes. Finally, \cref{fig:prompt3_negative_tool_free} illustrates the prompt designed for \textit{Negative Tool-Free Verify}, which forces the agent to identify and revise its initial hallucinations or reasoning flaws purely through visual self-reflection.

For the RL stage, to compute the IoU for the Grounding Reward, we require high-quality reference bounding boxes. These ground-truth coordinates are synthesized using Gemini-2.5-Pro. The detailed system prompt designed for this data annotation task is provided in \cref{fig:prompt4_rl_grounding}. During RL training of ReVISE, \cref{fig:prompt_revise_agent_rl} details the system prompt used to guide our ReVISE agent. This prompt explicitly delineates the agent's available action space (i.e., the suite of vision tools) and enforces a highly structured reasoning format (\texttt{<think>}, \texttt{<tool\_call>}, and \texttt{<answer>}). Crucially, to align with our proposed training objectives, the prompt integrates two specific behavioral constraints: (1) it explicitly instructs the model to actively verify intermediate visual evidence and refine tool parameters to foster self-correction; and (2) it requires the model to output the precise bounding boxes of relevant objects alongside the final answer, which serves as the necessary spatial output to compute the Grounding Reward ($r_{\text{grounding}}$) during RL updates. Notably, we prompt Gemini-2.5-Pro to output \emph{resolution-agnostic} normalized box coordinates on a $[0,1000]$ scale for consistent annotation across images. We then de-normalize them to absolute pixel coordinates in the original $(W\times H)$ space to compute IoU-based grounding rewards during RL.

\begin{figure*}[htbp]
  \centering
    \vspace{-14pt}
  \includegraphics[width=\linewidth]{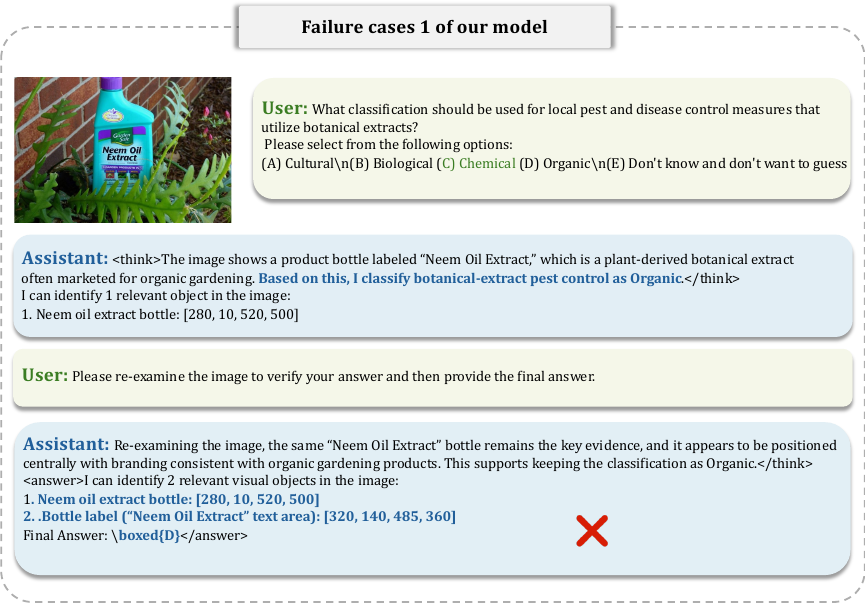}
  \caption{\textbf{Failure Case 1: Knowledge deficit and confirmation bias.} The model successfully grounds the visual evidence (reading the ``Neem Oil Extract'' label) but misclassifies it due to a lack of specialized agricultural knowledge. During the verification phase, re-examining the image merely reinforces its flawed heuristic, demonstrating that visual self-correction struggles to overcome fundamental gaps in abstract domain knowledge.}
  \label{fig:failure_case_1}
\end{figure*}
\begin{figure*}[t]
  \centering
  \includegraphics[width=\linewidth]{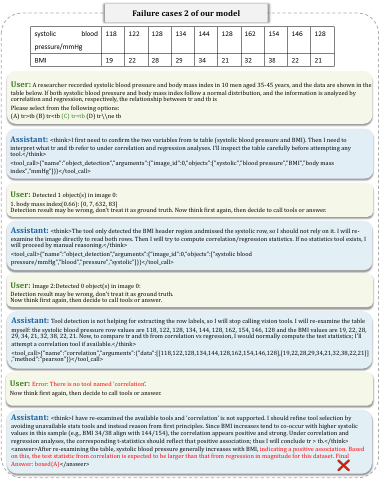}
  \caption{\textbf{Failure Case 2: Tool hallucination and theoretical reasoning limits.} While the model exhibits strong self-reflection by abandoning an ineffective \texttt{object\_detection} tool to manually parse the table, it ultimately fails. It hallucinates a non-existent \texttt{correlation} tool and relies on a flawed numerical heuristic, failing to recognize the mathematical equivalence of the test statistics ($t_r = t_b$).}
  \label{fig:failure_case_2}
\end{figure*}

\begin{figure*}[t]
  \centering
  \includegraphics[width=\linewidth]{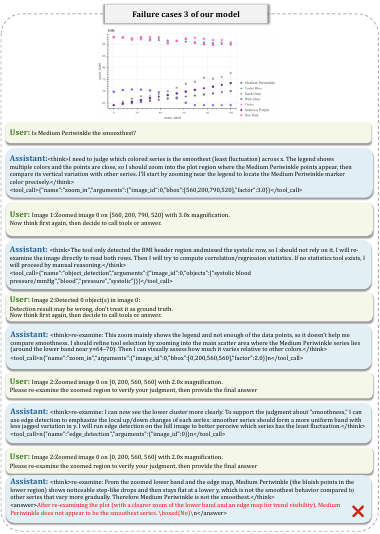}
  \caption{\textbf{Failure Case 3: Inappropriate Tool and Cumulative Errors.} In a cluttered spatial reasoning scenario, the model attempts to recover from noisy \texttt{object\_detection} outputs by invoking the \texttt{zoom\_in} tool. While this action provides localized clarity, it inadvertently strips away the global context. The agent consequently double-counts objects across crop boundaries and fails to reconcile the fragmented visual evidence.}
  \label{fig:failure_case_3}
\end{figure*}
\subsection{Dataset statistics}
For SFT, we leverage Gemini-2.5-pro to synthesize detailed tool-use trajectories and reasoning chains keeping only the instances where the model arrives at the correct final answer (Further details can be found in Appendix~\ref{subsec:system_prompt}).
Furthermore, we also generate ground-truth bounding boxes (for objects referenced in the question-answer pairs) using Gemini-2.5-pro (prompts shown in Appendix~\ref{subsec:system_prompt}) for the grounding reward in ~\cref{subsec:grpo}. These generated ground-truth bounding boxes are manually verified to ensure high-quality subset for RL training. As a result, we yield a high-quality cold-start dataset of 4k samples for SFT, alongside a robust set of 50k challenging QA pairs for RL training.

\section{Formulation of standard GRPO}\label{sec:grpo_fomulation}
\noindent \textbf{GRPO Formulation.} The model is treated as a policy $\pi_\theta$, which generates the output sequence $y$ conditioned on the input $(i, q)$. In particular, GRPO samples a group of $C$ candidate completions $y_1, y_2, \dots, y_C$ for each image-question pair under the current policy $\pi_\theta$. 
For each completion $y_k$, a scalar reward $r_k = R(i, q, y_k)$ is computed according to a task-specific reward function. Instead of relying on an explicit value estimator, GRPO computes a group-relative advantage by normalizing rewards within each sampled group:
\begin{equation}
\resizebox{0.5\linewidth}{!}{$
A_k =
\frac{
r_k - \mathrm{mean}\{r_1, \dots, r_C\}
}{
\mathrm{std}\{r_1, \dots, r_C\}
}.
$}
\end{equation}

\noindent \textbf{Optimization Objective.} We optimize $\theta$ by maximizing a GRPO-style clipped surrogate objective with group-normalized advantages:
\noindent \textbf{Optimization Objective.} We optimize $\theta$ by maximizing a GRPO-style clipped surrogate objective with group-normalized advantages:
\begin{equation}
\resizebox{0.89\linewidth}{!}{$
\begin{aligned}
\mathcal{J}_{\text{GRPO}}(\theta)
&= \frac{1}{G}\sum_{k=1}^{G} \Big[ \min\big(r_k A_k, \text{clip}(r_k,1-\epsilon,1+\epsilon)A_k\big) \\
&\quad -\beta\, D_{\mathrm{KL}}\big(\pi_\theta(\tau_k \mid q) \Vert \pi_{\mathrm{ref}}(\tau_k \mid q)\big) \Big]
\end{aligned}
\label{eq:rlobj}
$}
\end{equation}
where $G$ denotes the group size, specifically the number of distinct reasoning trajectories $\{\tau_k\}_{k=1}^{G}$ sampled per query $q$. $\pi_{\mathrm{ref}}$ is a fixed reference policy, $\beta$ controls the strength of KL regularization, and $\epsilon$ specifies the clipping range. The importance ratio is defined as $r_k = \frac{\pi_\theta(\tau_k \mid q)}{\pi_{\theta_{\mathrm{old}}}(\tau_k \mid q)}$, where $\theta_{\mathrm{old}}$ denotes the parameters used to generate the sampled outputs. This objective encourages reward improvement within each group while stabilizing updates around the reference policy.

\section{Additional Ablations and Analysis}
\label{sec:additional_ablations}

We provide additional analyses on the robustness and training design of ReVISE. 
Specifically, we study: (1) the sensitivity of GRPO training to noise in the grounding supervision, 
(2) whether mixing general-domain reasoning data mitigates the performance degradation on MMMU and MathVista, 
and (3) the role of cold-start SFT in establishing structured tool-use and self-correction behaviors.

\subsection{Robustness to Noisy Grounding Supervision}
\label{sec:box_noise}

Since the grounding reward used during GRPO relies on Gemini-generated bounding boxes, we first examine the quality of these boxes and evaluate the sensitivity of ReVISE to noisy spatial supervision.

We manually inspect 1,000 randomly sampled RL instances and find that approximately 98\% of the generated bounding boxes are correct, while only 2\% contain noticeable localization errors. This indicates that the grounding reward is generally based on reliable spatial supervision.

We further conduct a controlled noise-sensitivity experiment by perturbing the bounding boxes used during GRPO training. For a box with width $w$ and height $h$, each horizontal and vertical coordinate is perturbed according to
\begin{equation}
    x' = x + \sigma w \epsilon, \qquad
    y' = y + \sigma h \epsilon,
    \qquad \epsilon \sim \mathcal{N}(0,1),
\end{equation}
where $\sigma \in \{0.05, 0.10\}$ controls the noise magnitude.

\begin{table}[t]
\centering
\small

\resizebox{\columnwidth}{!}{
\begin{tabular}{lcccc}
\toprule
Setting & CountBench & CVBench & BLINK-HARD & Avg. \\
\midrule
w/o grounding reward
    & 73.68 & 81.15 & 61.20 & 72.01 \\
ReVISE-3B
    & \textbf{74.92} & \textbf{83.83} & \textbf{63.89} & \textbf{74.21} \\
5\% jitter
    & 74.30 {\scriptsize (-0.62)}
    & 83.08 {\scriptsize (-0.75)}
    & 63.37 {\scriptsize (-0.52)}
    & 73.58 {\scriptsize (-0.63)} \\
10\% jitter
    & 71.27 {\scriptsize (-3.65)}
    & 79.76 {\scriptsize (-4.07)}
    & 60.52 {\scriptsize (-3.37)}
    & 70.52 {\scriptsize (-3.69)} \\
\bottomrule
\end{tabular}
}
\caption{Sensitivity of ReVISE-3B to noise in the bounding boxes used for grounding supervision during GRPO training. Numbers in parentheses denote the absolute change relative to ReVISE-3B.}
\label{tab:box_noise}
\end{table}

As shown in Table~\ref{tab:box_noise}, introducing 5\% Gaussian jitter causes only a small average degradation of 0.63 points, suggesting that ReVISE is robust to moderate inaccuracies in the teacher-generated boxes. In contrast, increasing the perturbation to 10\% results in a substantially larger drop of 3.69 points on average. Nevertheless, the low empirical box-noise rate observed in our training data suggests that such severe perturbations are uncommon in practice.

GRPO clipping can further reduce the influence of occasional noisy supervision by limiting excessively large policy updates caused by high-variance rewards. However, clipping itself does not determine whether a bounding box is semantically correct and therefore cannot eliminate systematic teacher bias. In ReVISE, this risk is further mitigated by the low observed noise rate and the multi-component reward design, which combines grounding supervision with answer accuracy, format, and self-correction rewards.

\subsection{Effect of Mixed-Domain Training}
\label{sec:mixed_domain}

ReVISE primarily focuses on perception-heavy tasks that benefit from explicit visual grounding and tool-based verification. We observe that, while this specialization substantially improves CVBench, it leads to lower performance on general reasoning benchmarks such as MMMU and MathVista compared with the original Qwen2.5-VL-3B-Instruct model. We hypothesize that this degradation is mainly caused by the training-data distribution rather than an inherent conflict between self-correction and abstract reasoning.

To test this hypothesis, we augment both SFT and RL with math and science reasoning data. For SFT, we add 1K tool-free verification trajectories, consisting of 500 MAVIS-Function math samples and 500 ScienceQA samples. Gemini-2.5-Pro is used to construct trajectories containing explicit verification and self-correction steps. For RL, we additionally include 10K samples from the same domains with final-answer supervision.

Because object-level bounding boxes are not generally meaningful for these tasks, we adopt task-conditional reward computation. Specifically, the grounding reward is disabled for math and science samples by setting $\lambda_{\mathrm{ground}}=0$, while retaining answer-accuracy, format, and self-correction rewards. For the original perception-heavy samples, we keep $\lambda_{\mathrm{ground}}=1$.

\begin{table}[t]
\centering
\small
\resizebox{\columnwidth}{!}{
\begin{tabular}{lccc}
\toprule
Setting & CVBench & MMMU & MathVista \\
\midrule
Qwen2.5-VL-3B-Instruct
    & 72.55 & 49.00 & \textbf{61.20} \\
ReVISE-3B
    & 83.83 & 47.31 & 55.37 \\
ReVISE-3B + mixed-domain
    & \textbf{83.97} {\scriptsize (+0.14)}
    & \textbf{49.02} {\scriptsize (+1.71)}
    & 59.72 {\scriptsize (+4.35)} \\
\bottomrule
\end{tabular}}
\caption{Effect of mixed-domain training on perception-heavy and general reasoning benchmarks. Numbers in parentheses denote improvements over ReVISE-3B.}
\label{tab:mixed_domain}
\end{table}

Table~\ref{tab:mixed_domain} shows that mixed-domain training substantially recovers performance on MMMU and MathVista, improving them by 1.71 and 4.35 points, respectively, while preserving the strong CVBench performance of ReVISE. In particular, MMMU is recovered to approximately the performance of the original instruction-tuned model, while the CVBench gain remains intact. These results suggest that the degradation on general reasoning benchmarks mainly arises from domain specialization and can be mitigated by appropriately mixing general-domain reasoning data into training.

\subsection{Effect of Cold-Start SFT}
\label{sec:cold_start_sft}

Finally, we examine whether the proposed framework can be learned directly through RL without cold-start SFT. ReVISE requires the model to follow a structured multi-turn interaction protocol, including generating reasoning steps, producing valid tool calls, incorporating tool observations, and triggering self-correction when necessary. We therefore use SFT to initialize these behaviors before GRPO optimization.

\begin{table}[t]
\centering
\small

\begin{tabular}{lcc}
\toprule
Setting & CountBench & CVBench \\
\midrule
RL-only w/o cold-start SFT
    & 71.98 & 75.34 \\
ReVISE-3B w/o RL (SFT only)
    & 72.41 & 77.89 \\
ReVISE-3B (SFT + RL)
    & \textbf{74.92} & \textbf{83.83} \\
\bottomrule
\end{tabular}
\caption{Effect of cold-start SFT and RL on ReVISE-3B.}
\label{tab:sft_rl_ablation}
\end{table}

As shown in Table~\ref{tab:sft_rl_ablation}, directly applying RL without cold-start SFT performs substantially worse than the full ReVISE pipeline, particularly on CVBench. SFT alone establishes useful structured tool-use and verification behaviors, while subsequent RL further improves performance from 72.41 to 74.92 on CountBench and from 77.89 to 83.83 on CVBench.

These results indicate that the two stages play complementary roles: cold-start SFT provides a reliable initialization for structured tool interaction and self-correction, whereas RL improves the resulting policy once the model can consistently follow the required interaction protocol.

\section{Additional Qualitative Examples}\label{sec:add_qua_examples}
While ReVISE demonstrates strong capabilities in spatial reasoning and visual error recovery, it is important to understand its boundaries. In this section, we present representative failure cases to illustrate the challenges that persist, particularly in knowledge-intensive scenarios.

\noindent \textbf{Failure Case 1: Knowledge Deficit and Confirmation Bias.} \\
As shown in \cref{fig:failure_case_1}, the model is tasked with answering a scientific classification question regarding pest control measures based on an image of a ``Neem Oil Extract'' bottle. The correct agricultural classification for botanical extracts in this context is ``Chemical'' (biochemical pesticides). However, the model incorrectly answers ``Organic'' and fails to correct itself during the verification stage. 

\noindent \textbf{Failure Case 2: Tool Hallucination and Theoretical Knowledge Deficit.} \\
As illustrated in \cref{fig:failure_case_2}, the model encounters a mathematical statistics question involving a data table. The problem requires the theoretical knowledge that the $t$-statistic for testing a Pearson correlation coefficient is mathematically identical to the $t$-statistic for testing the slope in a simple linear regression (thus, $t_r = t_b$). 

\noindent \textbf{Failure Case 3: Inappropriate Tool and Cumulative Errors.} \\
As depicted in \cref{fig:failure_case_3}, the model tackles a complex spatial reasoning and counting task in a highly cluttered environment. It initiates the \texttt{object\_detection} tool, which returns noisy and heavily overlapping bounding boxes. Attempting to self-correct, the agent invokes \texttt{zoom\_in} on a localized region. However, by doing so, it loses the global spatial context, double-counts objects spanning across crop boundaries, and ultimately fails to reconcile the local zoomed evidence with the global scene.

We attribute the failures in both of these knowledge-intensive scenarios to two fundamental limitations of our current framework: \textbf{Misalignment Between Toolset and Abstract Task Requirements:} Our current suite of tools (e.g., object detection, zoom, depth estimation) is inherently designed for spatial, geometric, and perceptual reasoning. When faced with tasks requiring abstract knowledge or computation, the model either correctly avoids tool usage but hits a hard knowledge boundary (as in Case 1, where web search is needed), or it struggles to understand its action space and hallucinates non-existent tools (such as the \texttt{correlation} tool in Case 2). This highlights the necessity of expanding the tool library with knowledge retrieval and programmatic execution environments (e.g., a Python interpreter).

\textbf{Intrinsic Knowledge Gaps and Confirmation Bias:} In both cases, the model successfully grounds the visual evidence (reading the ``Neem Oil Extract'' label or extracting the data array from the table) but misinterprets the implications due to a lack of deep domain expertise (agricultural taxonomy or statistical theorems). Consequently, it falls back on flawed common-sense heuristics. Crucially, in these scenarios, our verification mechanism fails to correct the trajectory: prompting the model to \texttt{re-examine} the image merely reinforces its correct visual perception without injecting the missing abstract knowledge, inevitably leading to confirmation bias rather than successful self-correction.



\end{document}